\documentclass[10pt,twocolumn,letterpaper]{article}

\usepackage{cvpr}              % To produce the CAMERA-READY version
\usepackage{graphicx}
\usepackage{amsmath}
\usepackage{amssymb}
\usepackage{booktabs}
\usepackage{tabularx}
\usepackage{bm}

\usepackage[pagebackref,breaklinks,colorlinks]{hyperref}

\usepackage[capitalize]{cleveref}
\crefname{section}{Sec.}{Secs.}
\Crefname{section}{Section}{Sections}
\Crefname{table}{Table}{Tables}
\crefname{table}{Tab.}{Tabs.}

\def\cvprPaperID{0006} % *** Enter the CVPR Paper ID here
\def\confName{CVPR}
\def\confYear{2023}

\begin{document}

\title{Fully Automated Task Management for Generation, Execution, and Evaluation: A Framework for Fetch-and-Carry Tasks with Natural Language Instructions in Continuous Space}

\author{Motonari Kambara and Komei Sugiura \\
Keio University\\
{\tt\small \{motonari.k714, komei.sugiura\}@keio.jp}
% \and
% Komei Sugiura\\
% % First line of institution2 address\\
% {\tt\small komei.sugiura@keio.jp}
}
\maketitle

\vspace{-3mm}
%%%%%%%%% ABSTRACT
\begin{abstract}
\vspace{-4mm}
This paper aims to develop a framework that enables a robot to execute tasks based on visual information, in response to natural language instructions for Fetch-and-Carry with Object Grounding (FCOG) tasks. Although there have been many frameworks, they usually rely on manually given instruction sentences. Therefore, evaluations have only been conducted with fixed tasks. Furthermore, many multimodal language understanding models for the benchmarks only consider discrete actions. To address the limitations, we propose a framework for the full automation of the generation, execution, and evaluation of FCOG tasks. 
In addition, we introduce an approach to solving the FCOG tasks by dividing them into four distinct subtasks.
\end{abstract}
\vspace{-3mm}

\vspace{-3mm}
\section{Introduction}
\vspace{-2mm}
In our aging society, the need for daily care and support is increasing. As a result, the shortage of home care workers has become a social problem, and domestic service robots that can physically assist people with disabilities are attracting attention. However, the current ability of domestic support robots to understand instructions in natural language and execute daily tasks appropriately is insufficient.

The aim of this paper is to develop a framework for a robot to execute instructions based on visual information when given natural language instructions for fetch--and--carry tasks.
% 
% For example, consider an instruction sentence such as ``Go to the bedroom, grasp a rabbit doll, and place it on a corner sofa.'' In this case, it is desirable for the robot to navigate to the bedroom, grasp the rabbit doll, and then place it on a sofa located in the corner of the room.
% 
Many frameworks have been proposed for the execution of everyday tasks by robots based on natural language instructions \cite{shridhar2020alfred, shen2021igibson, teach, qi2020reverie, goyal2022ifor}. For example, ALFRED~\cite{shridhar2020alfred} is a large-scale benchmark containing 25K instructions. However, it is difficult to turn it into an on-the-fly simulation because the instructions are manually annotated.
Therefore, it is also difficult to evaluate with diverse tasks, and evaluations have only been conducted with fixed tasks.

Furthermore, many methods are also proposed in the benchmarks. In the ALFRED benchmark~\cite{shridhar2020alfred}, many multimodal language understanding models such as FILM~\cite{min2022film} and HLSM-MAT~\cite{ishikawa2022moment} have been proposed. Many of these multimodal language understanding models consider only discrete actions. Therefore, it is challenging to apply them to the real world, where continuous actions are required. On the other hand, the RoboCup@Home competition~\cite{iocchi2015robocup} is a representative framework for daily tasks by domestic service robots in the real world. In this framework, tasks are not generated automatically.

In this paper, we propose a framework for fully automating fetch-and-carry tasks in a simulation environment. This framework allows for the generation, execution, and evaluation of tasks on--the--fly. 
% The task generation system of this framework generates instruction sentences using a crossmodal instruction generation model.
% Therefore, it becomes possible to execute tasks based on instruction sentences of various forms. 
In the proposed framework, the simulation environment is a continuous space. 
% Therefore, robots are able to perform continuous actions.

% The main contributions of this paper are as follows:
% \vspace{-2mm}
% \begin{itemize}
%     \setlength{\parskip}{0.5mm} % 段落間
%     \setlength{\itemsep}{0.2mm} % 項目間
%     \item We propose a framework for fully automating the generation, execution, and evaluation of FCOG tasks. 
%     \item We propose an approach to solving the FCOG tasks by dividing them into four subtasks: Navigation, Object Location Retrieval (OLR), Fetching, and Carrying.
% \end{itemize}
% \vspace{-2mm}
\vspace{-3mm}
\section{Problem Statement}
\vspace{-2mm}
In this paper, we address the FCOG task. The task involves providing a robot with natural language instructions related to fetch--and--carry tasks, which the robot then executes utilizing visual information.

The inputs of the task are defined as natural language instruction and multiple images taken by a robot's camera. Given the input, the robot should grasp a target object and place it to a destination in accordance with the instruction.

\vspace{-3mm}
\section{Method}
\vspace{-2mm}
\begin{table*}[h]
    % \normalsize
    \vspace{-2mm}
    \caption{\small Quantitative comparison for FCOG tasks. The best scores are in bold.}
    \vspace{-2mm}
    \label{tab:quant_fcog}
    \centering
        % \begin{adjustbox}{width=\linewidth}
        \begin{tabular}{m{6cm}Wc{2.25cm}Wc{2.25cm}Wc{2.25cm}Wc{2.25cm}}
            \hline
            &  Navigation & OLR & Fetching & Carrying \\
            Methods & Success rate$[\%]$ &  Accuracy$[\%]$ & Success rate$[\%]$ & Success rate$[\%]$ \\
            \hline\hline
            Baseline (WRS-VS winning method) & $\bm{100}$ $(40/40)$ & $0$ $(0/0)$ & $0$ $(0/0)$ & $0$ $(0/0)$ \\
            % Ours (i) & $\bm{100} (40/40)$ & $0 (0/0)$ &  $0 (0/0)$ & $0 (0/0)$\\
            Ours & $\bm{100}$ $(40/40)$ & $\bm{20}$ $(8/40)$ &  $\bm{100}$ $(8/8)$ & $\bm{12.5}$ $(1/8)$\\
            \hline
        \end{tabular}
        % \end{adjustbox}
    \vspace{-5.5mm}
\end{table*}

The proposed framework is closely related to tasks like the General Purpose Service Robot task in RoboCup@Home competition~\cite{iocchi2015robocup}.
The framework fully automates the simulation of FCOG tasks using three systems: a task generation system, a task evaluation system, and a task execution system.

The Task Generator generates tasks through three main steps. First, the mechanism creates a simulated environment by placing static game objects. Next, it randomly places dynamic game objects in the environment and also positions the robot in a specified initial location for each environment. Finally, the task generation mechanism automatically generates tasks. In this step, the mechanism first randomly selects the objects to be used as target objects and destinations from the simulation environment.
The task generation mechanism then uses Unity functionality to obtain the coordinates of these objects and takes pictures of the objects using virtual cameras.
The Task Generator generates instruction sentences using the CRT~\cite{kambara2021case}. The model generates object manipulation instruction sentences based on the target region, destination region, and context regions.
% 
% 
% For the target region and destination regions, the generator generates them based on the segmentation images of the respective target objects and target regions. For the context regions, the generator uses the set of regions extracted by Faster R--CNN.
% 
% 
% For the example shown in Fig. Y, 
The model generates an instruction sentence like ``Move the wooden toy car in front of the white bottle onto the table.''
% Instruction sentences to navigate to the room where the target object is located are generated by rule--based methods. 
The Task Generator obtains the name of the room where the target object is located from the simulator, and uses it to generate an instruction sentence like ``Go to the bedroom.'' Finally, the Task Generator combines these instruction sentences.

In this study, we split the FCOG task into four subtasks: Navigation, Object Location Retrieval (OLR), Fetching, and Carrying.
The OLR can be roughly split into three steps: Crawling \& Image Collection, Object Detection and Multimodal Language Comprehension.

After identifying the target region and the destination region, the robot executes fetch--and--carry tasks. The Fetching task is carried out in the following specific steps. First, the robot moves to the point where it took the image. Then, it grasps the object using heuristic methods. Success in the Fetching task is defined as the ability to grasp the appropriate object.
% 
% If the Fetching task is successful, the robot performs the Carrying task in the following steps. 
% First, the robot moves to the point where it took the image. Then, it places the object on the furniture using heuristic methods. Success in the Carrying task is defined as the ability to place the appropriate object on the appropriate furniture.
If the Fetching task is successful, the robot performs the Carrying task in similar steps.

The Task Evaluator determines that the session has ended when it reaches the following states:
\vspace{-3mm}
\begin{itemize}
\setlength{\parskip}{0.2mm} % 段落間
\setlength{\itemsep}{0.2mm} % 項目間
    \item A certain amount of time has elapsed during the execution of the FCOG task.
    \item The robot successfully completes the FCOG task.
    \item The robot fails in executing any of the subtasks.
\end{itemize}
\vspace{-3mm}
After a session termination decision is made, the Task Generator initializes the environment and starts the next session.
% That is, it starts again from the step of placing static game objects.
% \input{pages/section_4.tex}
\vspace{-3mm}
\section{Experimental Results}
\vspace{-2mm}
We used an extended version of the standard simulator used in the World Robot Summit 2018 Partner Robot Challenge/Virtual Space Competition (WRS-VS,~\cite{okada2019competitions}).
% This simulator has a continuous action space, so the robot is capable of executing continuous actions.
% 
Table~\ref{tab:quant_fcog} shows the quantitative results for each subtask. The table displays the number of executions and successful attempts for each task.
We used the method~\cite{mizuchi2020optimization} that won WRS--VS as the baseline method.
In this method, trajectory generation for a robotic arm, path planning, and navigation methods were the same as those of the proposed method.
% From the table, it can be observed that all methods achieved a 100\% success rate in the Navigation task.

In the OLR task, the baseline method had an accuracy of 0\%. This is likely because it could not understand the diverse reference expressions contained in the instruction sentences generated by the Task Generator. 
On the other hand, the proposed method had an accuracy of 20\%.
\begin{figure}[t]
    \centering
    \includegraphics[height=75mm]{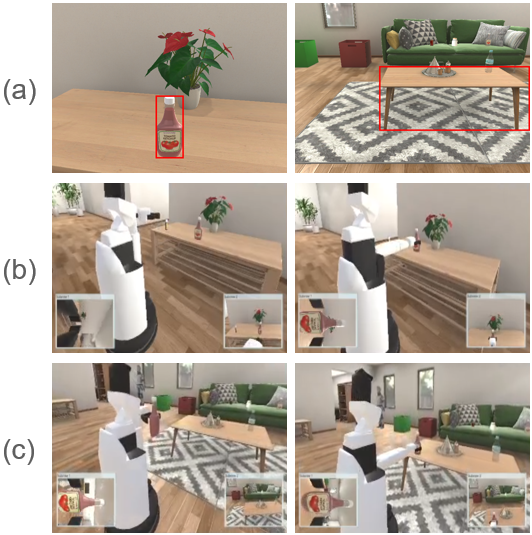}
    \vspace{-4mm}
    \caption{\small A successful session. In the figure, (a) shows the images of the target object and the destination acquired by the task generation system.}
    \vspace{-6mm}
    \label{fig:qua_fcog2}
\end{figure}
The baseline method did not execute the Fetching and Carrying tasks, as these tasks were only performed upon successful completion of the OLR task. In contrast, the proposed method achieved success rates of 20\% and 12.5\% for the Fetching and Carrying tasks, respectively.

% Furthermore, from the table, it can be seen that the accuracy of the OLR task in Condition (i) was 0\%. This suggests that using CLIP for reducing the number of images in the OLR task contributed to an improvement in accuracy.
% 
% To investigate whether reducing the number of collected images in the OLR task contributes to the improvement of accuracy, we established a condition (Condition (i)) in the ablation study, where the reduction using CLIP was not performed.

Fig.~\ref{fig:qua_fcog2} shows the qualitative results for the FCOG tasks.
% In each figure, (a) shows images of the target object and the destination obtained by the Task Generator. Here, the red rectangles are assigned based on the segmentation obtained from Unity. 
 % 
% Fig.~\ref{fig:qua_fcog2} shows a successful case.
% In the case shown in Fig. X, the Task Generator selected a box of cigarettes on a shelf and a red box as the target object and the destination. The instruction generated by the Task Generator was "Go to the bedroom, move a square object from the shelf to the side table.” 
% The robot successfully navigated to the only bedroom in the environment. Subsequently, in the OLR task, the robot appropriately identified a box of cigarettes on a shelf as the target object. Finally, as shown in Fig. X(b), the robot succeeded in grasping the box of cigarettes in the Fetching task.
% 
% 
% Similarly, in the case shown in Fig. Y, 
In this case, as shown in Fig.~\ref{fig:qua_fcog2}(a), the Task Generator selected a red bottle and a table in front of a sofa as the target object and the destination. The instruction generated by the Task Generator was "Go to the living room, move a plastic bottle from the shelf to the table.”
In the Navigation task, the robot successfully navigated to the living room. In the OLR task, a red bottle and a table in front of the desk were correctly identified as the target object and the destination, respectively. The robot then successfully grasped the red bottle in the Fetching task, as shown in Fig.~\ref{fig:qua_fcog2}(b). Finally, as shown in Fig.~\ref{fig:qua_fcog2}(c), the robot placed the red bottle on the table in the Carrying task.

\vspace{-3mm}
\section{Conclusions}
\vspace{-2mm}
In this paper, we focused on building a framework for FCOG tasks.
The main contributions are as follows:
\vspace{-3mm}
\begin{itemize}
    \setlength{\parskip}{0.2mm} % 段落間
    \setlength{\itemsep}{0.2mm} % 項目間
\item We proposed a framework for fully automating the generation, execution, and evaluation of FCOG tasks.
\item We proposed an approach to solving the FCOG tasks by dividing them into four subtasks.
% : Navigation, OLR, Fetching, and Carrying.
\end{itemize}
\vspace{-3mm}
% \noindent
\textbf{Acknowledgements} This work was partially supported by JSPS KAKENHI Grant Number JP23H03478, JST CREST, NEDO and JSPS Fellows Grant Number JP23KJ1917.

%%%%%%%%% REFERENCES
{\small
\bibliographystyle{ieee_fullname}
\bibliography{reference}

\begin{thebibliography}{10}\itemsep=-1pt

\bibitem{goyal2022ifor}
Ankit Goyal, Arsalan Mousavian, Chris Paxton, et~al.
\newblock {IFOR: Iterative Flow Minimization for Robotic Object Rearrangement}.
\newblock In {\em CVPR}, pages 14787--14797, 2022.

\bibitem{iocchi2015robocup}
Luca Iocchi, Dirk Holz, et~al.
\newblock {RoboCup@Home: Analysis and results of evolving competitions for
  domestic and service robots}.
\newblock {\em Artificial Intelligence}, 229:258--281, 2015.

\bibitem{ishikawa2022moment}
Shintaro Ishikawa and Komei Sugiura.
\newblock {Moment-based Adversarial Training for Embodied Language
  Comprehension}.
\newblock In {\em ICPR}, pages 4139--4145, 2022.

\bibitem{kambara2021case}
Motonari Kambara et~al.
\newblock {Case Relation Transformer: A Crossmodal Language Generation Model
  for Fetching Instructions}.
\newblock {\em IEEE RA-L}, 6(4):8371--8378, 2021.

\bibitem{min2022film}
So Min, Devendra Chaplot, Pradeep Ravikumar, Yonatan Bisk, et~al.
\newblock {FILM: Following Instructions in Language with Modular Methods}.
\newblock In {\em ICLR}, 2022.

\bibitem{mizuchi2020optimization}
Yoshiaki Mizuchi and Tetsunari Inamura.
\newblock {Optimization of Criterion for Objective Evaluation of HRI
  Performance that Approximates Subjective Evaluation: A Case Study in Robot
  Competition}.
\newblock {\em AR}, 34(3-4):142--156, 2020.

\bibitem{okada2019competitions}
Hiroyuki Okada, Tetsunari Inamura, and Kazuyoshi Wada.
\newblock {What competitions were conducted in the service categories of the
  World Robot Summit?}
\newblock {\em AR}, 33(17):900--910, 2019.

\bibitem{teach}
Aishwarya Padmakumar, Jesse Thomason, Ayush hrivastava, Patrick Lange, et~al.
\newblock {TEACh: Task-driven Embodied Agents that Chat}.
\newblock In {\em AAAI}, pages 2017--2025, 2022.

\bibitem{qi2020reverie}
Yuankai Qi, Qi Wu, Peter Anderson, et~al.
\newblock {REVERIE: Remote Embodied Visual Referring Expression in Real Indoor
  Environments}.
\newblock In {\em CVPR}, pages 9982--9991, 2020.

\bibitem{shen2021igibson}
Bokui Shen, Fei Xia, Chengshu Li, et~al.
\newblock {iGibson 1.0: A Simulation Environment for Interactive Tasks in Large
  Realistic Scenes}.
\newblock In {\em IROS}, pages 7520--7527, 2021.

\bibitem{shridhar2020alfred}
Mohit Shridhar, Jesse Thomason, Daniel Gordon, et~al.
\newblock {ALFRED: A Benchmark for Interpreting Grounded Instructions for
  Everyday Tasks}.
\newblock In {\em CVPR}, pages 10740--10749, 2020.

\end{thebibliography}
}

\end{document}